\documentclass{article}

\usepackage{PRIMEarxiv}

\usepackage[utf8]{inputenc} 
\usepackage[T1]{fontenc}    
\usepackage{textcomp}       
\usepackage{microtype}      

\usepackage{amsmath,amssymb,bm}

\usepackage{amsmath,amsfonts,bm}

\def\eqref#1{equation~\ref{#1}}

\def\1{\bm{1}}

\DeclareMathAlphabet{\mathsfit}{\encodingdefault}{\sfdefault}{m}{sl}
\SetMathAlphabet{\mathsfit}{bold}{\encodingdefault}{\sfdefault}{bx}{n}

\usepackage[table]{xcolor}  

\usepackage{booktabs}
\usepackage{makecell}
\usepackage{threeparttable}
\usepackage{multirow}
\usepackage{graphicx}
\usepackage{xspace}

\usepackage{natbib}
\setcitestyle{authoryear,round,citesep={;},aysep={,},yysep={;}}

\usepackage{url}
\usepackage{hyperref}
\hypersetup{
  colorlinks=true,
  linkcolor=blue,
  citecolor=blue,
  urlcolor=blue,
  filecolor=blue
}
\usepackage[capitalize]{cleveref}
  \makeatletter
  \renewcommand{\@toptitlebar}{%
    \vskip 0.1in
    \vskip -\parskip%
  }
  \renewcommand{\@bottomtitlebar}{%
    \vskip 0.05in
    \vskip -\parskip
    \vskip 0.05in%
  }
  \makeatother
\definecolor{tpgreen}{HTML}{2ECC40}
\definecolor{fnred}{HTML}{E02B1D}

\providecommand{\ours}{RECOUNT\xspace}

\title{RECOUNT: \underline{Re}ference-guided \underline{Coun}ting
with Synthetic \mbox{Visual} Exemplars
}

\author{
  Adriano D'Alessandro \\
  Simon Fraser University \\
  \texttt{acdaless@sfu.ca} \\
  \And
  Ali Mahdavi-Amiri \\
  Simon Fraser University \\
  \texttt{amahdavi@sfu.ca} \\
  \And
  Ghassan Hamarneh \\
  Simon Fraser University \\
  \texttt{hamarneh@sfu.ca} \\
}

\begin{document}
\maketitle

\begin{figure}[t]
  \centering
  \includegraphics[width=\textwidth]{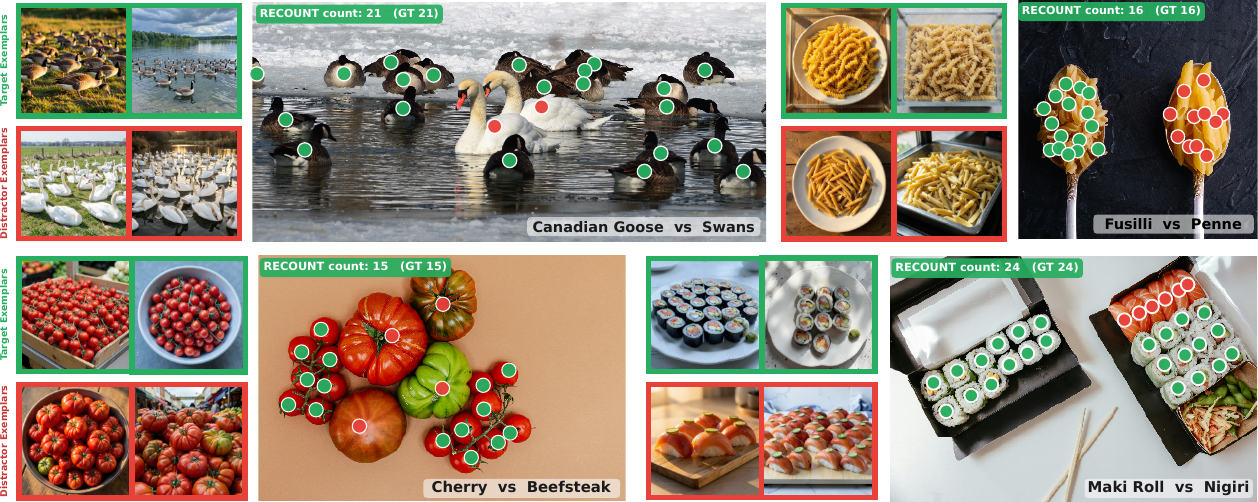}
  
  \caption{\textbf{From a reference image to precise counts in mixed category scenes.} We propose \ours{}, which automatically turns a reference image into an unlimited supply of \emph{synthetic visual exemplars} (SVEs). We use SVEs to create support sets that specify both what to count and what to reject. Applied as a lightweight filter over a frozen counter's point proposals, \ours{} pulls a single category out of multi-category scenes where language alone cannot. Green points are accepted, red points are rejected. Each panel pairs a query image with the target (top, green) and distractor (bottom, red) SVE galleries that drive the decision. Across these challenging scenes, \ours{} recovers the exact count.}
  \label{fig:teaser}
\end{figure}

\begin{abstract}
Text-guided zero-shot object counters excel at spatial localization but categorize poorly on novel or fine-grained classes: natural language is too coarse to fully specify visual identity, so they fail to separate visually similar distractors. Few-shot counters sidestep this with visual exemplars, but require manual annotations on every image. To resolve this dilemma, we introduce \ours{}, a plug-and-play framework for \emph{image-guided} zero-shot counting. Rather than specify a category with a text prompt, our key insight is to specify it visually, from a single off-scene reference image. However, we find that a lone reference image provides narrow coverage of a category's appearance and is unreliable across diverse scenes. We therefore repurpose a diffusion model as an automated contrastive data engine that expands the reference into a diverse exemplar gallery, supplying the discriminative detail that text cannot. \ours{} preserves the class-agnostic proposals of any frozen counter and offloads categorization to a separate visual module (a frozen backbone with a lightweight head trained on this synthetic data) that matches each proposal against the target and distractor galleries. Applied to a frozen counter, \ours{} attains the best zero-shot accuracy on both benchmarks, cutting counting error (MAE) by 55\% on \textsc{LookAlikes} and 21\% on PairTally relative to the strongest prior zero-shot counter.
\end{abstract}

\keywords{Object counting \and Zero-shot learning \and Synthetic data \and Diffusion models \and Visual exemplars}

\section{Introduction}\label{sec:intro}

\begin{figure}[t]
    \centering
    \includegraphics[width=\textwidth]{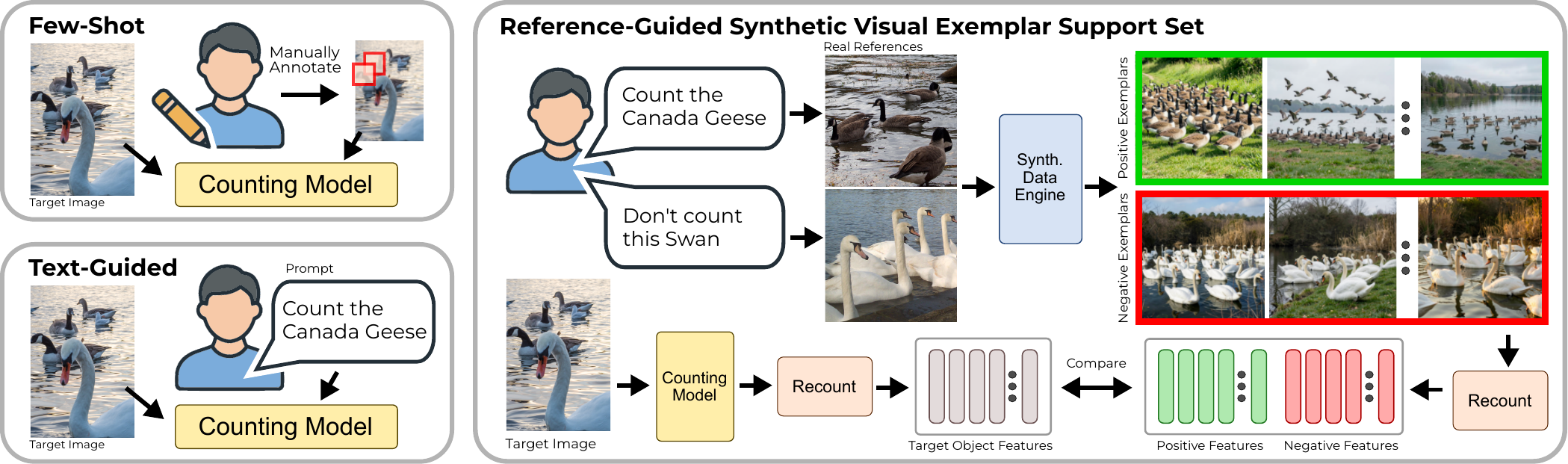}
    \caption{\textbf{A New Approach to Inference-Time Category Specification.} Existing class-agnostic counters require either per-image manual bounding-box annotations at test time (\emph{top-left}) or text prompts too coarse to specify visual identity (\emph{bottom-left}). In contrast, our framework (\emph{right}) specifies a category from a single real reference image, provided for both the target and its distractors. A synthetic data engine expands these lone references into a dense, visually diverse exemplar support set. Generated once per category, the support set is reusable across any number of query images. By deferring categorization until \emph{after} spatial localization, \ours{} matches each localized proposal against this support set, isolating the target from its distractors without per-image human labeling.}
    \label{fig:method_intro}
\end{figure}
Class-agnostic object counters have moved counting beyond fixed, closed-label regimes such as crowd counting~\citep{gao2020nwpu, lempitsky2010learning, chen2012feature, zhang2016single, sindagi2020jhu}, allowing a user to specify an arbitrary target category at inference time.  A category can be defined in one of two ways: with a text prompt, as in zero-shot counting, or with visual exemplars annotated on the query image, as in few-shot counting. In both cases, the counter extracts features from the reference and correlates them against the image, localizing every instance that matches the specification. This recipe has proven remarkably effective at spatial localization, reliably finding and separating objects across crowded and highly varied scenes.

However, this inference-time flexibility masks a common failure mode in open-world deployment: existing counters can identify \emph{where} an object is but they struggle to verify \emph{what} it is, often accepting visually similar distractors and background clutter as target instances. Two distinct limitations drive this failure. The first is that a text prompt cannot fully describe all visual attributes, and the cues that separate a Poblano pepper from a Jalape\~{n}o pepper are far easier to show visually than to put into words. The second is a data limitation, where counting datasets carry rich localization annotations but are starved of categories. FSC-147~\citep{Ranjan_2021_CVPR}, the standard training set for class-agnostic counting, provides thousands of dense, spatially annotated images but fewer than 100 training categories. This is ample supervision to learn object localization, yet far too narrow to learn dense categorization. Even when pretrained text encoders supply open-world vocabulary, a network trained on so few categories cannot ground fine textual distinctions, so the added semantics buy little. This aligns with broader findings that contrastive vision-language representations inherently struggle to ground fine-grained visual and spatial properties, a deficiency that persists regardless of data scale~\citep{Tong_2024_CVPR, yuksekgonul2023when}. The alternative, few-shot visual exemplars, avoids the pitfalls of language but demands a human annotation on every query image, forfeiting the scalable, annotation-free premise that makes zero-shot counting useful.

Our goal is to close the gap between these two regimes: to obtain the visual precision of few-shot exemplars at the annotation-free cost of a text prompt. Our key idea is to specify the target category with a reference image. To keep this scalable, the reference is provided once per category and reused across many query images, rather than annotated onto each query image as few-shot counting requires. A lone reference, however, captures the object from only one view in a single setting, too little to match the varied appearances an object takes across real scenes. We therefore re-introduce text, using a text-to-image diffusion model to expand the reference into a diverse gallery. Trained on billions of image-text pairs, such models can already place a novel object into new scenes: given the reference image and a set of text templates, they render the depicted object under diverse poses, lighting, and arrangements, turning a single image into a gallery of synthetic visual exemplars (SVEs). Here the templates control only the composition of the scene, while the object's identity comes entirely from the reference. We build one such gallery for the target and one for each of its \emph{distractors}, which are the visually similar categories it must be separated from, whether supplied by the user or identified automatically (e.g., by prompting an LLM for competing categories). The specification then captures both what to count and what to reject (Figure~\ref{fig:method_intro}). A single reference image thus yields a visual specification as expressive as a few-shot exemplar set yet as cheap as a text prompt, reusable across any image containing the target category. Because the category is specified by an image that is collected once and never annotated on the query, we call this paradigm \emph{image-guided} zero-shot counting.

We realize this as \ours{}, a plug-and-play module that attaches to a frozen counter, leaving its localization untouched and correcting only its categorization. For each candidate point the counter emits, \ours{} pools the dense, high-quality features of a frozen vision backbone at that location. The cues that separate a target from a distractor, however, are subtle and buried under coarse category-level similarity, so we pass each pooled feature through a lightweight projection head that surfaces them. We train this head with a contrastive objective on a large synthetic dataset built by the same engine that generates the exemplars, teaching it to discount the appearance categories share and amplify the fine differences that set a target apart from its distractors. At inference, we label each candidate by its nearest exemplar, accepting it only if it lies closer to a target exemplar than to any distractor. The generative engine thus serves the method twice over: offline it produces the training data that teaches the head to discriminate, and online it produces the reference-grounded exemplars that specify each new target, so adding a category never requires retraining.

\begin{itemize}
     \item \textbf{Image-guided zero-shot counting.} We introduce a new specification paradigm that defines a target category using a reference image, combining the visual specificity of exemplar-based counting with the annotation-free scalability of a text prompt. To make a single reference sufficient, we repurpose a text-to-image diffusion model as a contrastive data engine that expands it into diverse galleries of target and distractor exemplars, and use the same engine to build \textsc{SynthAlikes}, a synthetic corpus that supplies the category breadth counting data lacks.

    \item \textbf{\ours{}, a plug-and-play discrimination module.} We contribute a lightweight, model-agnostic filter that corrects a frozen counter's categorization while leaving its localization intact.

    \item \textbf{State-of-the-art in Fine-Grained Multi-Category Counting.} \ours{} sets a new zero-shot state of the art on \textsc{LookAlikes} and PairTally, cutting counting error by 55\% and 21\% over the strongest prior zero-shot counter, without annotating a single query image.
\end{itemize}
\section{Related Work}
\paragraph{Few-Shot Counting.} Few-shot counters specify the target with a small set of visual exemplars, typically bounding boxes drawn on the query image~\citep{liu2022countr,Dukic_2023_ICCV,Pelhan_2024_CVPR,countingdetr2022}. Because the specification is visual rather than textual, and matches the distribution of the query image, it captures object identity far more precisely than a language prompt. This precision, however, comes at the cost of a manual annotation on every query image, placing a human in the loop at inference and forfeiting the scalability of zero-shot counting. Our approach seeks to preserve the visual specificity of exemplar-based counting while removing this burden, synthesizing the exemplars from a single off-scene reference image which can be reused across every query image.

\paragraph{Zero-Shot Counting.} Zero-shot counting~\citep{va_count_zhu_2024,zhizhong2024point,Xu_2023_CVPR,kang2024vlcounter,zhangiccv2025} was formalized by \citet{Xu_2023_CVPR} (ZSC), which specifies the target category with a text prompt. Following the field's density-map tradition, early methods regressed a density map from this prompt. More recent work instead predicts explicit point or box locations, with several methods building on GroundingDINO~\citep{groundingdinoeccv}, including CountSE~\citep{liu2025countse}, CountGD~\citep{AminiNaieni24}, CountGD++~\citep{Amini-Naieni_2026_CVPR}, and GroundingREC~\citep{Dai_2024_CVPR}.

Most relevant to our work, several of these methods seek the precision of visual exemplars while remaining text-guided. ZSC~\citep{Xu_2023_CVPR}, CountSE~\citep{liu2025countse}, and CountGD++~\citep{Amini-Naieni_2026_CVPR} select representative visual features from the query image to act as pseudo-exemplars. This selection, however, is still driven by text and assumes the prompt can pick out the right features; in mixed-category scenes, where the target and its distractors co-occur, that assumption breaks down. Our approach removes text from the specification entirely: the exemplars are generated from an external reference image rather than selected from the query, so the target is defined by appearance instead of by a prompt applied to a cluttered scene. 

\paragraph{Counting with Synthetic Data.} Using synthetic data to support discrimination has become a prominent research direction~\citep{fake2real}. Early single-category counters benefited from 3D-rendered data such as the GCC dataset~\citep{wang2019learning}, while more recent counting works instead employ diffusion models. FiGO~\citep{dalessandro2025justsaytheword} generates data for the test-time adaptation of a segmentation model to add new categories, AFreeCA~\citep{d2024afreeca} exploits the weak quantity signal in generated images, and CountGD++~\citep{Amini-Naieni_2026_CVPR} generates a single canonical object from the query image to serve as a pseudo-exemplar. Our method differs from these in both the source and the use of its synthetic data. Rather than generating from text or from the query image, we ground generation on a single off-scene reference image and synthesize dense, multi-instance scenes that mirror deployment conditions. A single engine then drives both stages of our pipeline, training a lightweight discrimination head offline and specifying each new target online, while the base counter stays frozen, avoiding the per-category retraining used by test-time adaptation.

\section{\ours{}} \label{sec:method}
\begin{figure}[t]
    \centering
    \includegraphics[width=\textwidth]{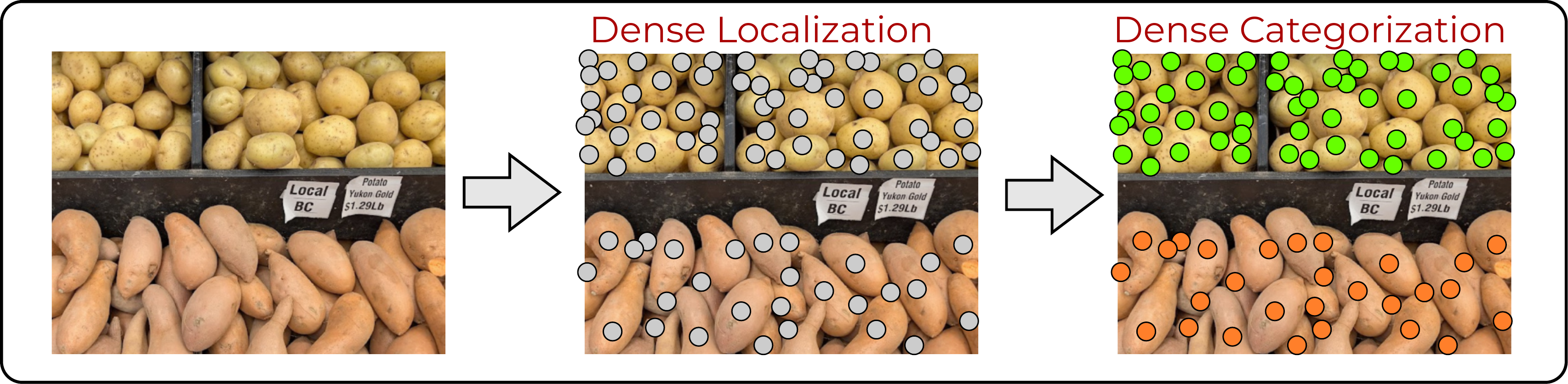}
    \caption{\textbf{Decoupling Dense Localization from Dense Categorization.} Object counting requires solving two distinct tasks: dense localization and dense categorization. Traditional models trained on single-category benchmarks (e.g., FSC147) conflate these steps; they excel at localizing objects (\textit{center}) but fail to separate visually similar or novel categories. Decoupling these operations treats localization as a class-agnostic geometry problem, leaving semantic boundaries to be resolved independently (\textit{right}).}

    \label{fig:dense_categorization}
\end{figure}
Counting an arbitrary category decomposes into two tasks: localizing the objects (dense localization) and deciding which of them belong to the target (dense categorization). Counting is simply their composition (see: Fig~\ref{fig:dense_categorization}). As established in Section~\ref{sec:intro}, localization is already data-rich, with datasets like FSC-147~\citep{Ranjan_2021_CVPR} providing thousands of images with dozens of location-based annotations per image. However, categorization is the bottleneck. We therefore leave localization to a frozen counter and focus on the problem of dense categorization. This requires answering two questions: how to \emph{specify} a target at inference without annotating the query image, and how to \emph{learn} to recognize it without dense, category-labeled data. \ours{} answers both with a single generative data engine.

Online, the data engine \emph{specifies} a category: from one real reference image, it synthesizes a visual support set, a gallery of target exemplars alongside galleries of its distractors, and we categorize each localized point by matching it against them. Offline, the same engine supplies the missing \emph{training} signal: conditioning FLUX.2~\citep{flux-2-2025} on reference images drawn from public repositories such as iNaturalist and Pexels, we generate a corpus of dense scenes spanning over a thousand categories, the taxonomic breadth that datasets like FSC-147~\citep{Ranjan_2021_CVPR} lack. The rest of this section details how we generate these scenes and support sets (Section~\ref{sec:synthesis}) and how the discrimination module learns to categorize against them (Section~\ref{sec:projection}).

\subsection{A Synthetic Data Engine for Dense Categorization}
\label{sec:synthesis}
\subsubsection{Reference-Guided Image Generation}
\begin{figure}[t]
    \centering
    \includegraphics[width=\textwidth]{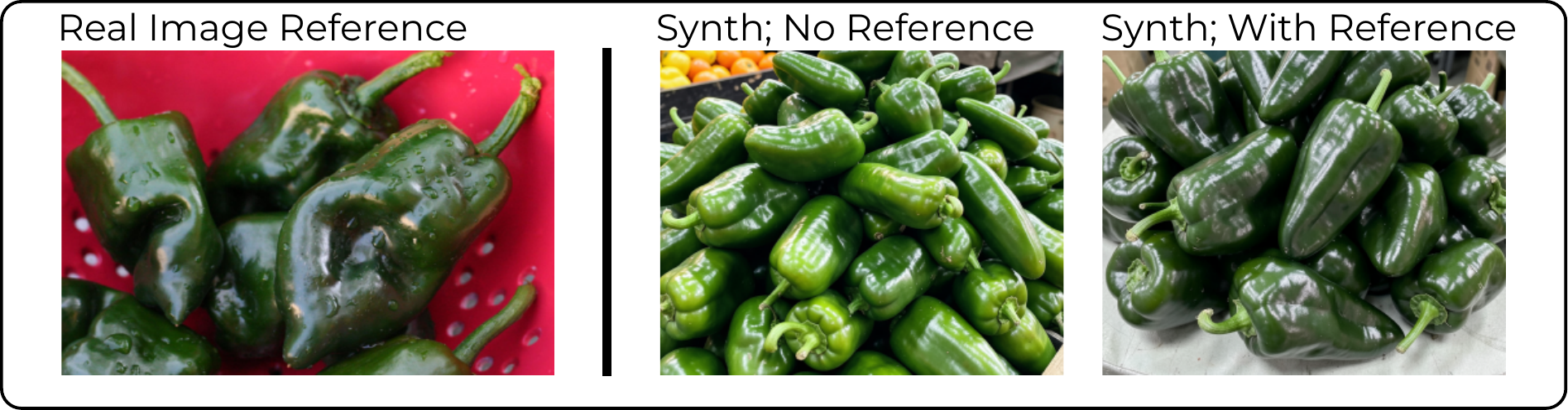}
    \caption{\textbf{Fine-Grained Data Generation with Image References.} Standard text-to-image models struggle to synthesize exact fine-grained details from text prompts alone. Given the prompt ``Poblano peppers,'' and a dense description, a text-only generation (\textit{center}) defaults to producing generic green pepper traits. By conditioning on a single real reference image (\textit{left}), we can design synthetic training data that capture domain-specific fine-grained visual features (\textit{right}).}
    \label{fig:using_a_reference}
\end{figure}
To answer the ``what'' question of dense categorization, we build a synthetic data engine that produces an unlimited supply of crowded, single-category images that accurately depict a category's specific visual attributes. However, as we argued earlier, the obstacle is that text alone cannot specify that appearance. Figure~\ref{fig:using_a_reference} shows why: prompted with ``poblano peppers,'' a diffusion model (FLUX.2) falls back on a generic green pepper, because a category name compresses a rich visual identity into a few tokens. Conditioned on a single real reference image, the same model instead preserves the true identity of the poblano. Our engine is built on the premise that identity comes from a reference image rather than a prompt.

A reference alone, though, only specifies the category's identity independent of how it appears in a full scene. For that, we use prompt templates. Diffusion models like FLUX.2 are trained on billions of images and carry broad knowledge of how objects populate the real world: how waterfowl gather on a pond, how produce is stacked at a market stall, etc. Our templates draw on this knowledge, placing the reference's precise visual identity into diverse, crowded scenes with varied viewpoints, arrangements, and lighting.
\begin{figure}[t]
    \centering
    \includegraphics[width=\textwidth]{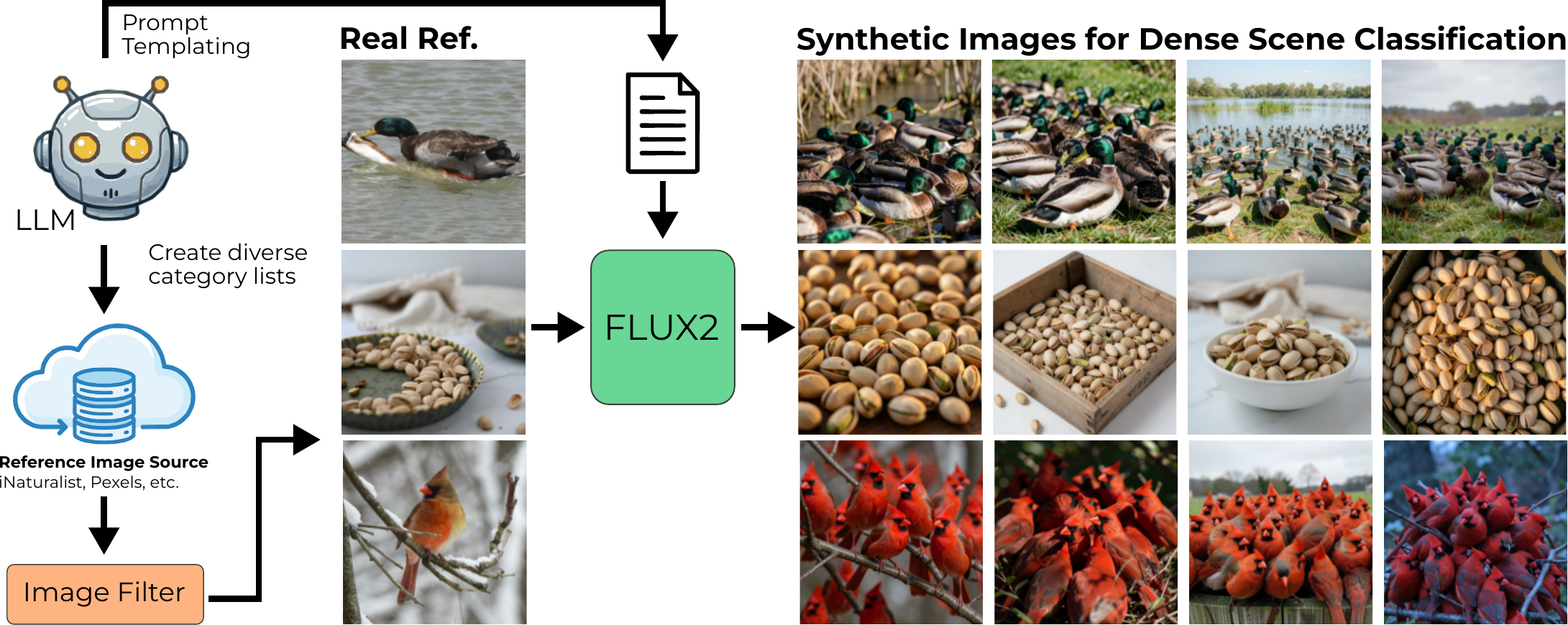}
    \caption{\textbf{Automated Generation of a Fine-Grained Dense Categorization Training Dataset.} Our data engine synthesizes dense training scenes by coupling real-world taxonomic relationships with generative image expansion. We initialize our categories by combining the iNaturalist taxonomy with manually seeded non-natural object classes. An LLM structurally parses this input to construct a massive, unified fine-grained taxonomy, which dictates image-fetching queries across diverse online repositories. The collected real-world reference images are filtered and then paired with LLM-generated scene templates, guiding a latent diffusion model to synthesize dense, visually complex scenes with accurate fine-grained object categories.}
    \label{fig:data_generation}
\end{figure}
\subsubsection{The \textsc{SynthAlikes} Dataset}
\label{sec:synthalikes}

With the data engine defined, we return to the question of how we can train a discriminative model to recognize a specific target category, and reject the categories that most resemble it, without annotating new dense data. We propose \textsc{SynthAlikes}, a scalable synthetic training set for dense categorization in which, for every category, we define the lookalike distractors that the model must reject.

\begin{figure}[t]
    \centering
    \includegraphics[width=\textwidth]{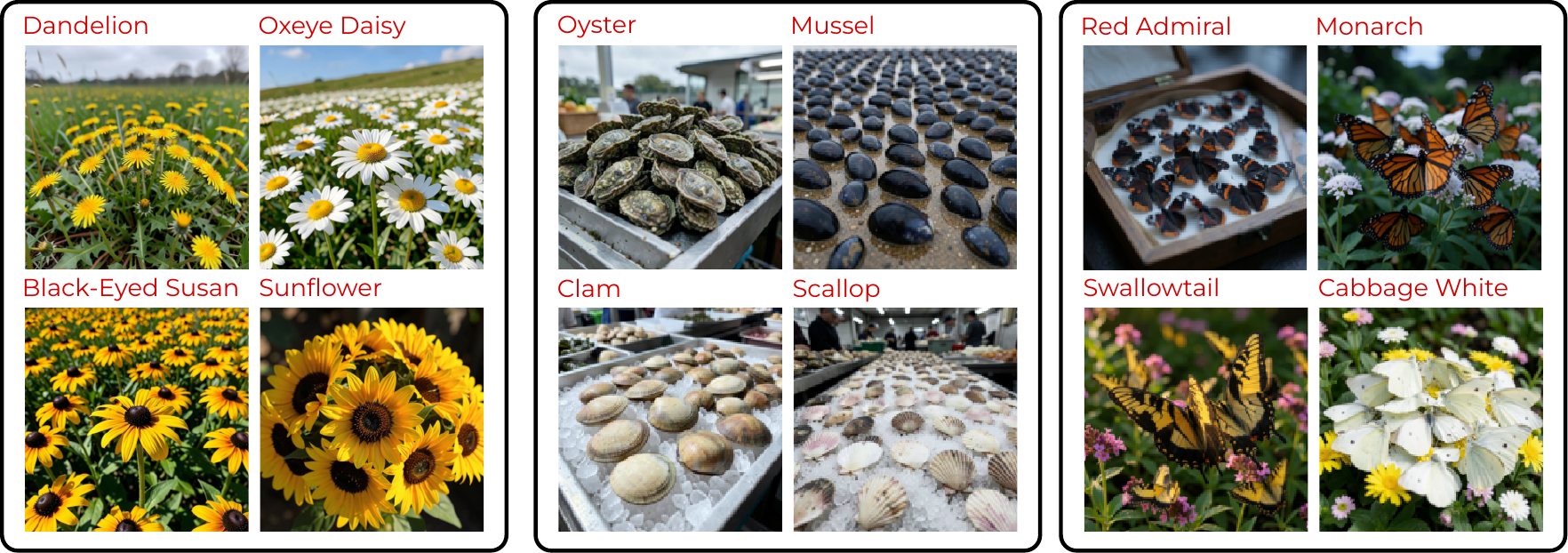}
    \caption{\textbf{Generated Fine-Grained Training Groups.} 
    Sample outputs from our automated data engine showing three distinct semantic groups (\textit{Asteraceae} flower species, shellfish, butterfly species) containing visually similar but distinct siblings. By organizing dense, multi-object scenes into contrastive sibling subsets, this data provides the explicit training signal required for our downstream dense categorization module.}
    \label{fig:synthalikes_groups}
\end{figure}

The construction of \textsc{SynthAlikes} proceeds in three stages. First, we establish a structured, fine-grained category taxonomy: using an LLM, we merge the iNaturalist hierarchy with a set of LLM curated classes (e.g., coin denominations, fastener types, citrus cultivars) into a unified taxonomy spanning over a thousand categories. Second, for each category, we retrieve candidate reference images from open repositories (e.g., iNaturalist, Pexels) and apply an automated quality filter, using CLIP to keep only clear depictions that match the broad category. We then populate the dataset by conditioning the data engine on these references and a set of structured prompt templates, synthesizing diverse, dense, single-category scenes (Figure~\ref{fig:data_generation}). Third, we obtain contrastive supervision directly from the taxonomy. Categories that sit together as taxonomic siblings, such as the flower species, shellfish, and butterflies in
Figure~\ref{fig:synthalikes_groups}, are precisely the categories most easily confused with one another, so we treat each category's siblings as its hard negatives. 

Finally, we pseudo-label every scene with a frozen class-agnostic counter. Because each scene contains a single known category, generating annotations for dense categorization does not require perfect localization. Instead, is simply requires that a sufficient number of the category's instances are approximately marked, which an off-the-shelf counter can provide. Each resulting synthetic image has a fine-grained category name, a set of points localizing at least one target instance, and a set of visually related hard negative siblings.

\subsubsection{Synthetic Visual Exemplars} \label{sec:sve_explained}
We now return to the question of how we specify, at inference time, which category to count and which to reject, without annotating the target image (i.e., few-shot counting). We answer it with the same engine, run online: from a single reference image per category, we synthesize a \emph{support set} of synthetic visual exemplars (SVEs). This provides a gallery of target SVEs alongside its visually similar distractor categories, and specifies at inference exactly what to tell apart.

Concretely, a support set has two sides: a positive gallery of SVEs depicting the target, and a negative gallery of SVEs for each of its distractors. The distractors are drawn the same way they were during training, as the target's siblings in the taxonomy or as distractors enumerated with the query, so a category's inference-time negatives are the very distractors it learned to reject. We use a gallery rather than the reference alone because the engine's SVEs span the target's appearance, its poses, lighting, and arrangements, so that an object in the scene need only resemble one exemplar to be recognized as the target, whereas a lone reference captures a single view and generalizes poorly. The support set is therefore the complete specification a frozen model needs to count a new category: it requires no manual annotations on the target image, it is built once per category and reused across all of that category's images, and it leaves the model itself unchanged. How the support set is used to produce a count is the subject of Section~\ref{sec:inference}.

\begin{figure}[t]
    \centering
    \includegraphics[width=\textwidth]{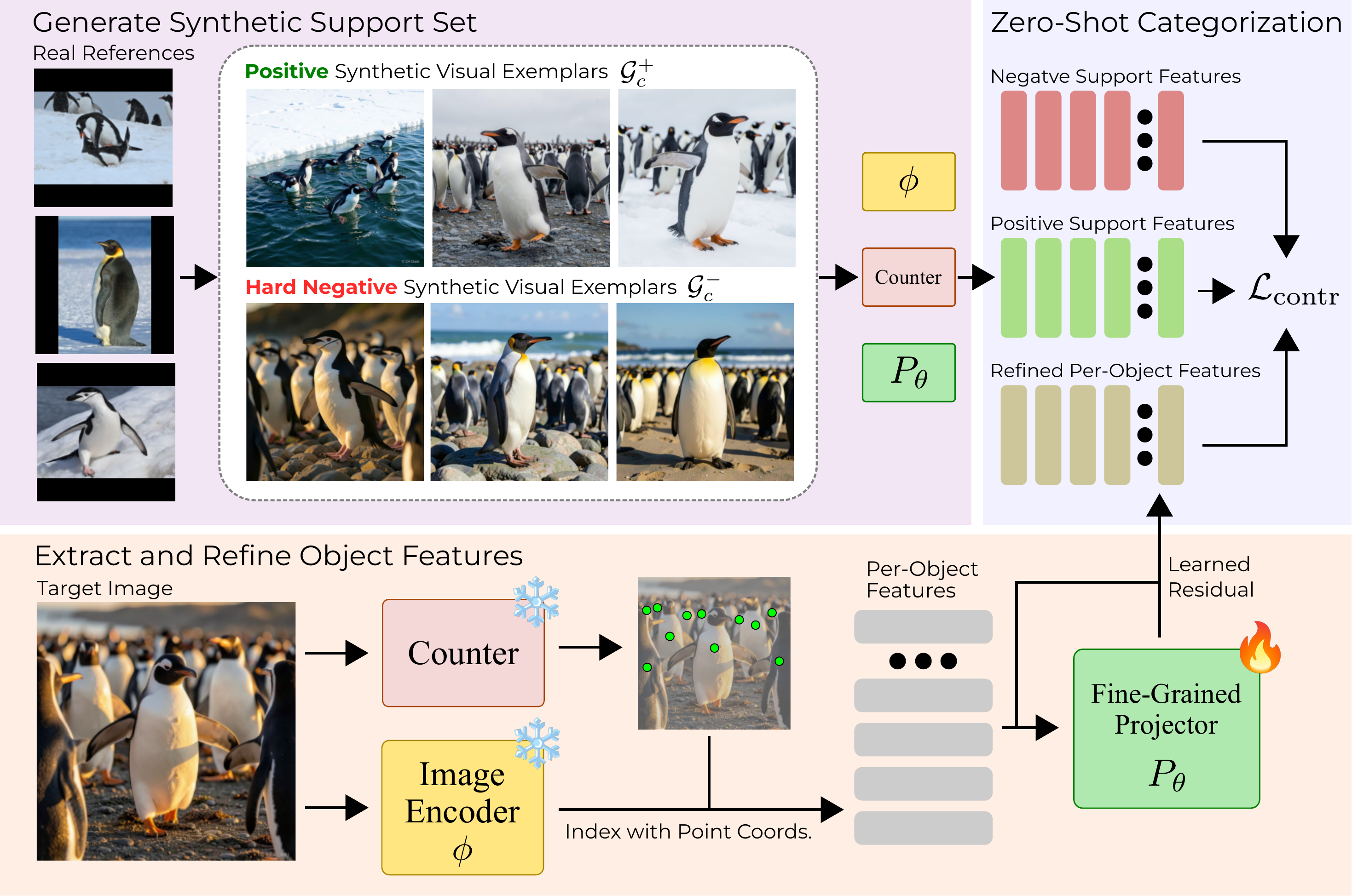}
    \caption{\textbf{Contrastive Training Pipeline.} We train the projection head $P_\theta$ on synthetic dense scenes, leaving the backbone and counter frozen. (Top)~From a few real references per category, our engine generates a synthetic support set for a category $c$: positive exemplars $\mathcal{G}^+_c$ depicting the target, and hard-negative exemplars $\mathcal{G}^-_c$ depicting its visually similar distractors. A frozen image encoder $\phi$, a frozen counter, and the projector $P_\theta$ reduce every exemplar to a support feature. (Bottom)~For a synthetic scene of $c$, the frozen counter pseudo-labels each instance with a point and $\phi$ produces a feature grid; indexing the grid at these points yields per-object features, which $P_\theta$ refines through a learned residual. (Right)~A multi-negative contrastive loss $\mathcal{L}_\mathrm{contr}$ pulls each refined per-object feature toward the positive support features and away from the negative support features. Only $P_\theta$ is trained.}
    \label{fig:main_methodology}
\end{figure}
\begin{figure}[t]
    \centering
    \includegraphics[width=\textwidth]{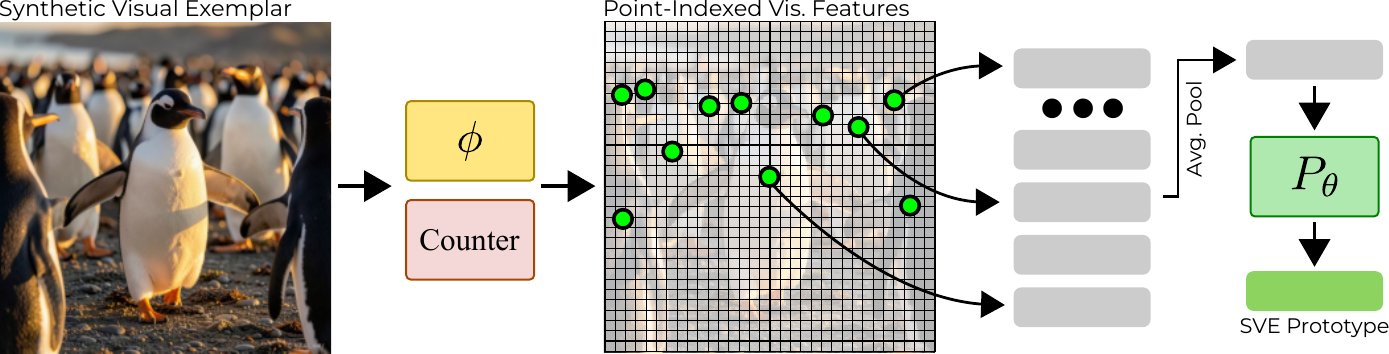}
    \caption{\textbf{Point Pooling for Synthetic Visual Exemplars.} Each synthetic visual
      exemplar (SVE) in a support gallery is reduced to a single prototype. The frozen counter
      proposes a point on every instance in the exemplar; we sample the backbone ($\phi$) feature
      grid at these points by bilinear interpolation, average the sampled features into one
      descriptor, and project it with the head $P_\theta$ to obtain the exemplar's prototype. A
      support gallery is the set of these prototypes, one per exemplar.}
    \label{fig:point_pool}
\end{figure}
\subsection{Surfacing Discriminative Visual Features}
\label{sec:projection}
Matching a candidate point against the support set is only meaningful in a feature space where distractor categories are separable. Modern self-supervised vision backbones already encode the visual detail needed to tell them apart~\citep{Tong_2024_CVPR}, but we find that these details are faint: they tend to sit beneath a dominant signal that captures only an object's broad category, so a raw cosine similarity collapses visually similar distractors together. We recover it with a lightweight projection head that reshapes the space so similarity reflects an object's specific identity rather than its coarse category.

Let $\phi$ be the frozen backbone. For each candidate point $p$ proposed by the counter,
we pool $\phi$'s feature grid at $p$ into a descriptor $f_p \in \mathbb{R}^d$. A
projection head $P_\theta$ maps every descriptor to a unit embedding,
\begin{equation}
P_\theta(f) = \frac{f + g_\theta(f)}{\lVert f + g_\theta(f) \rVert_2},
\end{equation}
where $g_\theta$ is a small two-layer MLP. The residual form keeps the embedding close to
the backbone's own geometry, perturbing it just enough to expose the discriminative
features without discarding what the backbone already represents. The head is
\emph{category-agnostic}: it has no per-category parameters and is applied identically to
candidate descriptors and to the SVE exemplars that form the support set. It is a purifier
of a shared feature space rather than a classifier, which is what lets a single head serve
every category, and transfer across backbones, without retraining.
\subsubsection{Contrastive Training}
\label{sec:training}
We train $P_\theta$ on \textsc{SynthAlikes} with a multi-negative InfoNCE objective
(Figure~\ref{fig:main_methodology}). Each step draws a synthetic scene of a category $c$.
The frozen counter pseudo-labels the scene, and we pool the backbone features at each labeled
point and project them, giving a set of per-object embeddings $\{P_\theta(f_p)\}$. Their
positives are the projected exemplars of $c$'s own support gallery $\mathcal{G}^+_c$; their
negatives $\mathcal{G}^-_c$ are the exemplars of $c$'s distractors, drawn from its siblings in
the taxonomy and augmented with a few cross-group negatives. The multi-negative InfoNCE loss
then pulls each embedding toward the positive exemplars and away from every negative in a
single softmax,
\begin{equation}
\mathcal{L}_\mathrm{contr} = -\sum_{p} \log
\frac{\sum_{g \in \mathcal{G}^+_c} \exp\!\big(\langle P_\theta(f_p), P_\theta(g)\rangle/\tau\big)}
     {\sum_{g \in \mathcal{G}^+_c \cup \mathcal{G}^-_c} \exp\!\big(\langle P_\theta(f_p), P_\theta(g)\rangle/\tau\big)},
\end{equation}
with temperature $\tau$ and $\langle\cdot,\cdot\rangle$ the cosine similarity between unit embeddings. Because the negatives are $c$'s own distractors, the loss gives $P_\theta$ gradient precisely where the backbone is ambiguous, forcing it to surface the features that separate confusable categories rather than the coarse category signal it already exposes. We
additionally apply a small dispersion regularizer that penalizes the pairwise similarity among $c$'s positive prototypes, preventing them from collapsing to a single embedding. Only $P_\theta$ is trained; the counter and backbone stay frozen.

\subsection{Inference}
\label{sec:inference}
At inference we are given a target category, its support set, and a test image. The frozen
counter proposes a set of candidate points, one per detected object, and our task is to
decide which of them belong to the target. Following Section~\ref{sec:projection}, we pool
and project each candidate into an embedding $P_\theta(f_p)$, and we reduce the support set
to a positive gallery $\mathcal{G}^+$ of target prototypes and a negative gallery
$\mathcal{G}^-$ of distractor prototypes (one prototype per exemplar,
Figure~\ref{fig:point_pool}). A candidate is accepted when it more closely resembles the
target than any distractor,
\begin{equation}
\mathrm{accept}(p) = \big[\, s^+(p) > s^-(p) \,\big], \qquad
s^+(p) = \max_{g \in \mathcal{G}^+} \langle P_\theta(f_p), P_\theta(g)\rangle, \quad
s^-(p) = \max_{g \in \mathcal{G}^-} \langle P_\theta(f_p), P_\theta(g)\rangle,
\end{equation}
and the predicted count is the number of accepted candidates. This is a nearest-exemplar rule: the
single closest prototype across both galleries decides each candidate's label, with no
per-category threshold to tune. Specifying a new target only swaps $\mathcal{G}^+$ and
$\mathcal{G}^-$, leaving the counter, the backbone, and $P_\theta$ untouched. We call $m(p) = s^+(p) - s^-(p)$ the margin of a candidate and the rule above accepts at $m(p) > 0$.

\begin{figure}[b]
    \centering
    \includegraphics[width=\textwidth]{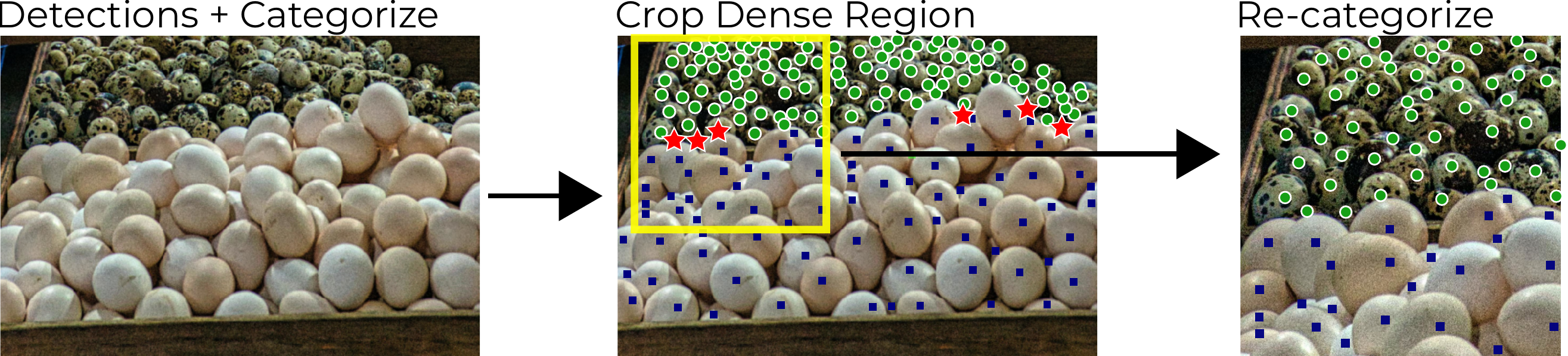}
    \caption{\textbf{Crop-and-Reprocess.} The pre-trained vision backbones that we utilize emit a fixed $32\times32$ patch grid, so in dense regions many instances occupy a single patch and cannot be meaningfully separated, causing the filter to miss or misjudge them (\emph{left}). We detect each dense cluster, crop it (\emph{middle}), and re-extract features at native resolution before re-categorizing the enclosed points (\emph{right}), recovering instances the base grid could not resolve.}
    \label{fig:crop_and_resize}
\end{figure}
\subsubsection{Partitioning Dense Images}
\label{sec:crop_and_reprocess}
The decision of Section~\ref{sec:inference} treats each candidate independently, which is reliable only when its descriptor $f_p$ isolates a single object. Common vision backbones emit a fixed feature grid, e.g. $32\times32$ for a $512$-pixel input, so in the densest regions several instances fall within one grid cell and are pooled into a shared descriptor. Their features blend, and a distractor packed among targets inherits the target's appearance, degrading precisely the candidates that are most crowded.

We resolve these regions with \emph{crop-and-reprocess}. We detect grid cells that contain more than one candidate, crop a padded box around each such dense cluster from the native-resolution image, and re-extract features from the crop in isolation, so that every enclosed candidate is pooled from its own portion of a fresh grid (Figure~\ref{fig:crop_and_resize}). A candidate in a dense cluster then carries two descriptors, one from the base grid and one from its crop, each yielding a margin $m(p)$. We keep the decision made with the larger $|m(p)|$, trusting whichever scale is more confident, while candidates in sparse regions retain their base-grid decision unchanged. The reprocessing is thus confined to the crowded regions that need it, recovering instances the base grid cannot resolve at negligible cost elsewhere.
\section{Experiments}

\subsection{Implementation Details}
\label{sec:implementation}

\paragraph{Backbones and Proposal Generation.} 
All primary experiments utilize a frozen DINOv3 ViT-B/16 backbone. Input images are resized to $512{\times}512$ and passed through the backbone to extract a $32{\times}32$ patch-token grid ($16$-px patches, $768$-d features). For cross-backbone validation, we additionally evaluate frozen DINOv2-B/14 (input $448{\times}448$, $32{\times}32$ grid) and FLAIR backbones; separate heads are trained for each backbone architecture. Point proposals are generated by a frozen CountGD counter~\citep{AminiNaieni24} pre-trained on FSC-147, prompted with broad category text and using its default preprocessing (shortest side $800$, max side $1333$). This proposal generator is applied identically across all evaluated methods; only the downstream filtering stage differs.

\paragraph{Synthetic Visual Exemplar (SVE) Galleries.} 
To construct category galleries, we synthesize $k{=}12$ Synthetic Visual Exemplars (SVEs) per category using FLUX.2-klein-4B~\citep{flux-2-2025} ($4$ denoising steps at $1024{\times}1024$ resolution). Generation is conditioned on a single external reference image per category retrieved off-scene from the web (never the query image) and prompted using diverse crowd-composition templates completed once per category via an external LLM. Each category is generated from a template ``\emph{A photograph of \{object\} \{setting\}, \{arrangement|crowd-composition\}. \{camera\}, \{lighting\}},'' appended with a fixed exclusivity clause that enforces a single object type and individually countable instances. Beyond the object name, we draw additional scene descriptions from an LLM generated dictionary of archetypes that cover broad categories like birds, produce, office supplies, etc. These additional descriptions supply type-appropriate arrangements, crowd compositions, camera framings, and lighting. Negative galleries follow the identical generation pipeline over a curated set of relevant distractors. 

Following CountGD++~\citep{Amini-Naieni_2026_CVPR}, we adopt the provided positive and negative class lists for each dataset. For \textsc{LookAlikes}, ground-truth annotations label all instances matching the broad category but group unnamed objects into an \textit{other} bucket. To address this, we prompt an LLM to generate a small set of co-occurring distractor categories (e.g., pairing \textit{Goldfields} as a distractor for \textit{California Poppies}). When benchmarking competing methods on \textsc{LookAlikes}, we use this identical set of positive and negative classes across all baselines.

\paragraph{Feature Extraction and Projection Head.} 
Each SVE image is processed through CountGD and the frozen vision backbone. Foreground point features are extracted via bilinear interpolation over the patch grid and averaged across detected instances, producing one $768$-d prototype per SVE image. A category gallery consists of the resulting set of $12$ prototypes. Query image points are extracted analogously via bilinear sampling at the proposed coordinate.

The only trainable component is a lightweight, category-agnostic residual projection head: a $2$-layer MLP ($768{\to}768$) with dropout ($0.1$), residual skip connections, and $\ell_2$-normalized outputs. This head is applied identically to query point features and gallery prototypes, functioning as a category-agnostic feature purifier rather than a parameterized classifier.

\paragraph{Training and Optimization.} 
The projection head is trained using a multi-negative InfoNCE objective (temperature $\tau{=}0.07$) on our \textsc{SynthAlikes} dataset (Section~\ref{sec:synthesis}). For each query prototype, the positive target is its corresponding gallery, while the denominator comprises all same-group sibling galleries plus $n{=}2$ sampled cross-group negatives, regularized by a dispersion loss (weight $0.02$). We optimize using AdamW (learning rate $5{\times}10^{-5}$, weight decay $10^{-4}$, batch size $8$) for $1$ epoch on a single NVIDIA H100 GPU over the \textsc{SynthAlikes} point annotations. Because the vision backbone and CountGD counter remain frozen throughout, only the $\sim$1.2M parameters of the residual projection head are updated.

\paragraph{Benchmarking Metrics \& Protocols.} For our evaluations on the \textsc{LookAlikes} and PairTally datasets, we use mean absolute error (MAE), and root mean squared error (RMSE), which both provide a measure of the difference between the ground truth count and the predicted count for each image. On \textsc{LookAlikes}, we follow the protocol of~\citep{dalessandro2025justsaytheword}: MAE and RMSE are computed per fine-grained subcategory and macro-averaged over all 37 subcategories to deal with data imbalance, so that no categories dominate the average. For all datasets, we use the crop-and-reprocess strategy detailed in Section~\ref{sec:crop_and_reprocess}.
\paragraph{Ablation Metrics \& Protocols.} To isolate dense categorization performance from localization quality, our ablations operate on the \emph{ground-truth} point localizations of the \textsc{LookAlikes} dataset: because each image is fully annotated for its target subcategory and its co-present lookalikes, we can score category assignment directly, independent of the performance of any counter. As described in Section~\ref{sec:inference}, a point is accepted as the target iff $m(p) > \delta$. We report Recall and the False Positive Rate (FPR) at the operating threshold $\delta{=}0$: Recall is the fraction of true query points accepted, and FPR is the fraction of lookalike points erroneously accepted as the target. To summarize separability independently of any single threshold, we report \textbf{AUC}, the area under the ROC curve traced by sweeping $\delta$. This is the probability that a randomly chosen query point receives a higher margin than a randomly chosen lookalike point. AUC thus measures how well the representation separates a target from a lookalike, while Recall and FPR characterize the deployed $\delta{=}0$ operating point. Since our ablation experiments are primarily concerned with feature quality, we perform them without the crop-and-reprocess strategy, unless stated otherwise.
\subsection{Results}
\paragraph{Cross-Dataset Generalization}
\begin{table*}[t]
  \centering
  \begin{threeparttable}
\caption{%
  \textbf{Cross-domain Transfer Performance Under Varying Inference Protocols.} We evaluate cross-domain transfer from the single-category FSC-147 dataset (A)~\citep{Ranjan_2021_CVPR} to two fine-grained, multi-category counting datasets: \textsc{LookAlikes} (B)~\citep{dalessandro2025justsaytheword} and PairTally (C)~\citep{nguyen2025pairtally}. Notation $\text{A}\rightarrow\text{B/C}$ denotes the training set (left) and test set (right). \textbf{Reference modalities:} $\oplus$ and $\ominus$ indicate the inclusion of positive and negative categories, respectively; BBox and Image represent bounding boxes from the target image and separate independent reference images; and \textsc{+pseudo} denotes text-selected pseudo-bounding boxes. Bold indicates the best score for each inference mode, and underline marks the overall best result. Note that \textsc{LookAlikes} is zero-shot only and cannot be evaluated in few-shot mode. Gray shading highlights our method.
}
  \label{tab:domain_adaptation_clean}
  \setlength{\tabcolsep}{6pt}
  \begin{tabular}{l l c  cc cc}
    \toprule
    \multirow{2}{*}{Method} & \multirow{2}{*}{Reference} & \multirow{2}{*}{Inference} & \multicolumn{2}{c}{$\text{A}\rightarrow\text{B}$} & \multicolumn{2}{c}{$\text{A}\rightarrow\text{C}$} \\
    \cmidrule(lr){4-5} \cmidrule(lr){6-7} 
    & &  & MAE$\downarrow$ & RMSE$\downarrow$ & MAE$\downarrow$ & RMSE$\downarrow$ \\
    \midrule
    PSeCo \footnotesize{+FiGO}~\citeyearpar{dalessandro2025justsaytheword} & Text & TTA &  19.57 & 30.86 & - & - \\
    DAVE \footnotesize{+FiGO}~\citeyearpar{dalessandro2025justsaytheword} & Text & TTA &  21.19 & 31.53 & - & - \\
    CountGD \footnotesize{+FiGO}~\citeyearpar{dalessandro2025justsaytheword} & Text & TTA &  \textbf{13.31} & \textbf{23.36} & - & - \\
    GrREC \footnotesize{+FiGO}~\citeyearpar{dalessandro2025justsaytheword} & Text & TTA &  16.15 & 26.99 & - & - \\
    \midrule
    CountGD\footnotesize{}~\citeyearpar{AminiNaieni24} & BBox & Few-Shot &  - & - & 46.67 & 70.85 \\
    DAVE\footnotesize{}~\citeyearpar{Pelhan_2024_CVPR} & BBox & Few-Shot &  - & - & 47.37 & - \\
    CountGD++\footnotesize{}~\citeyearpar{Amini-Naieni_2026_CVPR} & BBox$_\oplus$ & Few-Shot &  - & - & 46.41 & 69.52 \\
    CountGD++\footnotesize{}~\citeyearpar{Amini-Naieni_2026_CVPR} & BBox$^\ominus_\oplus$ & Few-Shot &  - & - & 35.27 & \underline{\textbf{60.85}} \\
    \rowcolor{gray!10} CountGD \footnotesize{+\ours{}} & BBox$_{\oplus}^\ominus$ & Few-Shot &  - & - & \underline{\textbf{34.90}} & 63.50 \\
    \midrule

    PSeCo\footnotesize{}~\citeyearpar{zhizhong2024point} & Text & Zero-Shot &  42.88 & 56.67 & - & - \\
    DAVE\footnotesize{}~\citeyearpar{Pelhan_2024_CVPR} & Text & Zero-Shot &  53.10 & 64.44 & - & - \\
    CountGD\footnotesize{}~\citeyearpar{AminiNaieni24} & Text & Zero-Shot &  30.17 & 45.01 & 50.32 & - \\
    GrREC\footnotesize{}~\citeyearpar{Dai_2024_CVPR} & Text & Zero-Shot &  18.51 & 30.39 & - & - \\
    CountGD++\footnotesize{}~\citeyearpar{Amini-Naieni_2026_CVPR} & Text$^\ominus_\oplus$ & Zero-Shot &  24.6\textsuperscript{\textdagger} & 37.3\textsuperscript{\textdagger} & 66.9\textsuperscript{\textdagger} & 102.4\textsuperscript{\textdagger} \\
    CountGD++\footnotesize{}~\citeyearpar{Amini-Naieni_2026_CVPR} & Text$^\ominus_\oplus$\tiny{\textsc{+pseudo}} & Zero-Shot &  29.1\textsuperscript{\textdagger} & 45.0\textsuperscript{\textdagger} & 65.5\textsuperscript{\textdagger} & 98.6\textsuperscript{\textdagger} \\

    \rowcolor{gray!10} CountGD \footnotesize{+\ours{}}  & Image$_{\oplus}^\ominus$ & Zero-Shot &  \underline{\textbf{8.41}} & \underline{\textbf{14.67}} & \textbf{39.96} & \textbf{65.63} \\
    \bottomrule
 
  \end{tabular}
  \begin{tablenotes}
      \small
      \item[$\dagger$] ~ Our evaluation using the official released implementations. 
    \end{tablenotes}
  \end{threeparttable}
\end{table*}
In Table~\ref{tab:domain_adaptation_clean}, we evaluate CountGD as a base model with the addition of \ours{} and compare performance in the domain adaptation setting, transferring from the FSC147 dataset~\citep{Ranjan_2021_CVPR} to either the \textsc{LookAlikes}~\citep{dalessandro2025justsaytheword} or PairTally~\citep{nguyen2025pairtally} datasets. We evaluate \ours{} as an image-guided zero-shot method, given that category specification at inference time requires only a single reference image per category, and that single reference is reusable and valid for any arbitrary query image. Our image-guided zero-shot strategy is also different from prior few-shot counting strategies, which typically require human-annotated exemplars on the query image, whereas \ours{} does not require any manual images and is not dependent on the query image. 

Compared to other zero-shot methods, we find that the addition of \ours{} to CountGD leads to substantial performance improvements on the PairTally and \textsc{LookAlikes} test sets. Specifically, \ours{} reduces PairTally test-set MAE from 50.32 to 39.96 (where lower is better), representing a 20.59\% performance improvement over the baseline zero-shot CountGD. We also find that it leads to a 72.12\% performance improvement on \textsc{LookAlikes} test-set MAE over the baseline zero-shot CountGD and a 54.6\% reduction over GroundingREC, the best performing text-guided zero-shot method. Our method also provides a 36.81\% improvement over zero-shot CountGD modified with FiGO, which is the best test-time adaptation strategy. We also find that in the few-shot setting, where we apply \ours{} to CountGD with in-domain visual exemplars, we achieve a 25.22\% improvement over CountGD and equivalent performance to CountGD++.

Further, we perform our own evaluation of CountGD++~\citep{Amini-Naieni_2026_CVPR}, benchmarking it on the \textsc{LookAlikes} dataset and PairTally dataset in zero-shot mode, using both positive and negative text, and pseudo exemplar settings. In both cases, we find that \ours{} not only leads to a performance improvement over the CountGD baseline, but that it leads to CountGD outperforming CountGD++ without any additional training. This suggests that existing zero-shot object counting methods like CountGD are already quite good at dense object localization, but have substantial room for improvement on dense categorization. Also, interestingly, CountGD++ using pseudo-labels underperforms compared to text alone, suggesting that the text may be selecting incorrect pseudo-exemplars, leading to a performance degradation. 

\paragraph{Category Adherence}
\begin{table*}[t]
  \centering
\caption{%
    \textbf{Prompt Adherence Results.} We benchmark our method against other zero-shot methods using the PrACo dataset~\citep{ciampi2025mind}, which measures whether a model correctly counts the \emph{prompted} category. The \emph{Negative Test} prompts each image with every class and reports Normalized Mean of Negative (NMN), which is the mean of the count predicted for \emph{absent} classes, normalized by the ground-truth count of the present class, and Positive Class Count Nearness (PCCN), which is the percentage of images whose positive-class count is closer to ground truth than its negative-class counts. The \emph{Mosaic Test} stacks the target image with a distractor-class image and counts the target over the collage, reporting counting precision (CntP) and recall (CntR); \emph{Text} = text-grounded cues (and, for CountGD++, text-synthesized exemplars); \emph{+pseudo} = pseudo-exemplars detected on the query image; \emph{Image} = a reference image that was not derived from the query image. Best per column in \textbf{bold}.
  }
  \label{tab:prompt_adherence}
  \setlength{\tabcolsep}{6pt}
  \begin{tabular}{l l cc cc}
    \toprule
    \multirow{2}{*}{Method} & \multirow{2}{*}{Reference} & \multicolumn{2}{c}{Negative Test} & \multicolumn{2}{c}{Mosaic test} \\
    \cmidrule(lr){3-4} \cmidrule(lr){5-6} 
    & & NMN$\downarrow$ & PCCN$\uparrow$ & CntP$\uparrow$ & CntR$\uparrow$ \\
    \midrule
    TFPOC~\citeyearpar{shi2024training} & Text$_\oplus$ & 0.75 & 66.04 & 0.69 & 0.85 \\
    DAVE~\citeyearpar{Pelhan_2024_CVPR} & Text$_\oplus$ & 1.05 & 37.02 & 0.84 & 0.80 \\
    DAVE~\citeyearpar{Pelhan_2024_CVPR} & Text$^\ominus_\oplus$ & 0.08 & 97.62 & 0.84 & 0.80 \\
    CountGD++~\citeyearpar{Amini-Naieni_2026_CVPR} & Text$_\oplus$\tiny{+pseudo} & 0.88 & 62.86 & 0.86 & 0.96 \\
    CountGD++~\citeyearpar{Amini-Naieni_2026_CVPR} & Text$^\ominus_\oplus$\tiny{+pseudo} &  0.07 & 97.99 & \textbf{0.90} & 0.96 \\
    \rowcolor{gray!10} CountGD \footnotesize{+\ours{}} & Image$_{\oplus}^\ominus$ &  \textbf{0.02} & \textbf{98.15} & \textbf{0.90} & \textbf{0.97} \\
    \bottomrule
  \end{tabular}
\end{table*}
\begin{figure}[p]
    \centering
    \includegraphics[width=0.95\textwidth]{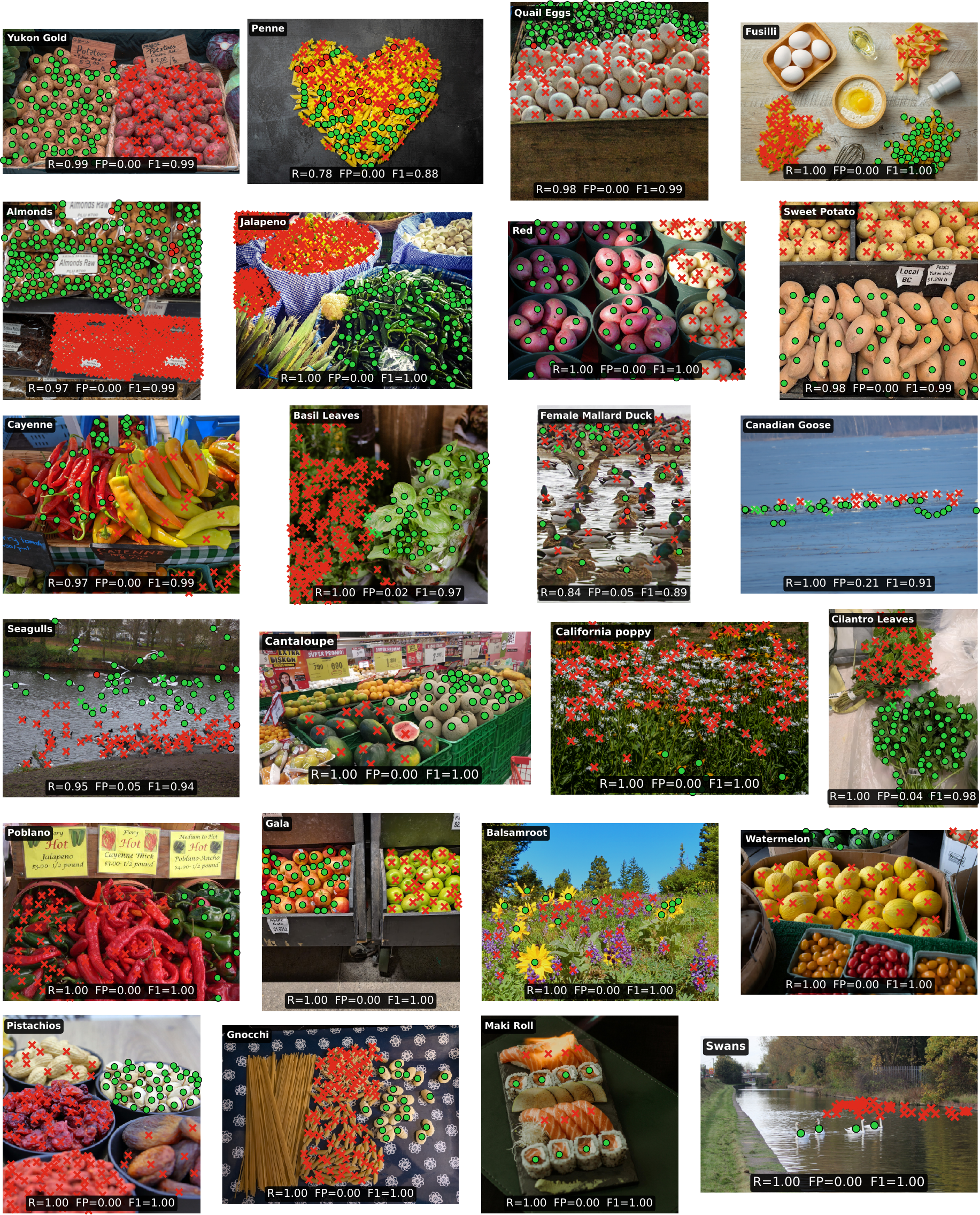}
    \caption{\textbf{Qualitative results on \textsc{LookAlikes}.} Each panel shows our filter's per-point decisions for a target category in a scene containing co-present lookalikes. In this \emph{oracle} setting, the filter is applied to ground-truth points to isolate categorization from detection. Markers encode decisions: a \textcolor{tpgreen}{\textbf{green circle}} is a target point \emph{accepted} (true positive), a \textcolor{fnred}{\textbf{red circle}} is a target point \emph{rejected} (false negative), a \textcolor{fnred}{\textbf{red $\times$}} is a distractor point \emph{rejected} (true negative), and a \textcolor{tpgreen}{\textbf{green $\times$}} is a distractor point \emph{accepted} (false positive). We report recall (R), false-positive rate (FP), and F1 at operating threshold $\delta{=}0$. Examples span 24 fine-grained categories across \textsc{LookAlikes}, showing the filter accepts the target while rejecting near-identical siblings.}
    \label{fig:qualitative_oracle}
\end{figure}

In Table~\ref{tab:prompt_adherence}, we evaluate the performance of our method on the PrACo benchmark~\citep{ciampi2025mind}, which measures whether a model adheres to the specified category---a known failure mode for many zero-shot counters. Category adherence is evaluated using two metrics. The \emph{Negative Test} prompts each image with every candidate category and reports NMN (the mean count predicted for absent categories normalized by the ground-truth count of the present one) and PCCN (the percentage of images whose present-category count is closer to ground truth than its absent-category counts; higher is better). The \emph{Mosaic Test} composites a query image with a distractor-category image and reports counting precision (CntP) and recall (CntR), penalizing objects counted in the wrong-category region.

\ours{} achieves top performance across every criterion. On the Negative Test, our method achieves an NMN of 0.02 and a PCCN of 98.15, demonstrating that it suppresses counts on incorrect categories in the vast majority of cases. On the Mosaic Test, our method, when combined with CountGD, yields performance equivalent to CountGD++. This suggests that CountGD and CountGD++ possess similar capabilities regarding dense object localization, with dense categorization being their primary point of differentiation.

\paragraph{Qualitative Results} Figure~\ref{fig:qualitative_oracle} illustrates our filter's per-point decisions across 26 fine-grained categories from \textsc{LookAlikes}, applied to ground-truth localizations so that the visualization isolates categorization from detection. In each scene, our SVE gallery accepts instances of the specified category while rejecting co-present visually similar distractors. \ours{} succeeds on genuinely challenging fine-grained distinctions---separating poblano from jalape\~no, and cayenne peppers; almonds from pistachios; and cilantro leaves from basil leaves. Our method does so even when the target and its distractors are interleaved in the same region, where per-point appearance is most obfuscated. The reported per-image recall, false-positive rate, and F1 confirm that in most scenes the filter cleanly partitions the target cultivar or species from its neighbors.

\subsection{The Impacts of Synthetic Data}
In this section, we isolate the specific contribution of the synthetic visual exemplars and the generation pipeline that we used for both the \textsc{SynthAlikes} training dataset and the SVE galleries. As described in Section~\ref{sec:implementation}, we use the ground-truth localization points provided by the \textsc{LookAlikes} dataset as an oracle so that we can quantify recall, FPR, and AUC, entirely decoupled from localization performance.

\paragraph{SVE Gallery}

\providecommand{\ph}{--}

\begin{table}[t]
  \centering
  \caption{%
    \textbf{Ablation over architectural components and reference modalities.}
    Evaluating the impact of vision backbone architecture, the \ours{} head, and reference modality on performance. Evaluated using the \textsc{Lookalikes} dataset with oracle localizations.
  }
  \label{tab:backbone_head_ablation}
  \setlength{\tabcolsep}{7pt}
  \begin{tabular}{ccc ccc}
    \toprule
    \multirow{2}{*}{Backbone} & \multirow{2}{*}{Head} & \multirow{2}{*}{Reference} & \multicolumn{3}{c}{Metrics} \\
    \cmidrule(lr){4-6}
    & & & Recall$\uparrow$ & FPR$\downarrow$ & AUC$\uparrow$ \\
    \midrule
    DINOv3 & dino.txt & Text      & 0.535 & 0.086 & 0.804 \\
    DINOv3 & None & Reference Only     & 0.624 & \textbf{0.083} & 0.809 \\
    DINOv3   & None   & SVE Gallery  & 0.849 & 0.116 & 0.913 \\
    DINOv3   & \ours{}   & SVE Gallery & \textbf{0.870} & 0.096 & \textbf{0.929} \\
    \midrule
    DINOv2 & None & Reference Only     & 0.542 & \textbf{0.065} & 0.765 \\
    DINOv2    & None   & SVE Gallery  & 0.814 & 0.116 & 0.884\\
    DINOv2   & \ours{}   & SVE Gallery  & \textbf{0.861} & 0.091 & \textbf{0.916} \\
    \midrule
    FLAIR & None & Text     & 0.279 & 0.156 & 0.653 \\
    FLAIR & None & Reference Only     & 0.323 & \textbf{0.132} & 0.644 \\
    FLAIR    & None   & SVE Gallery  & 0.651 & 0.293 & 0.755 \\
    FLAIR   & \ours{}   & SVE Gallery  & \textbf{0.707} & 0.250 &  \textbf{0.769} \\
    \bottomrule
  \end{tabular}
\end{table}
In Table~\ref{tab:backbone_head_ablation}, we isolate the performance improvement attributed to the \ours{} head (as described in Section~\ref{sec:projection}) and the SVE gallery across several baseline foundation vision models to demonstrate that it provides a real tangible benefit that cannot be attributed to a single underlying model. We specifically evaluate DINOv2~\citep{oquab2023dinov2, jose2025dinov2, ICLR2024_0b408293}, DINOv3~\citep{simeoni2025dinov3}, and FLAIR~\citep{xiao2025flair}, which are all foundation vision or vision-language models that focus on generating high-quality, highly descriptive, and dense images features. When we match vision features between the target object and reference images or SVE gallery, we use cosine similarity to measure the similarity.

For the DINOv3 backbone, we first evaluate dino.txt~\citep{jose2025dinov2} as a baseline for open-vocabulary zero-shot inference. The authors train dino.txt using CommonCrawl, a dataset with 2.3 billion image-text pairs. We find that \ours{} provides an absolute 33.5-point increase in recall, and 12.5-point gain in AUC, despite being trained on 0.00001\% as much data. This highlights that task specific, reference grounded synthetic data can provide significantly better results than broad and exceptionally large datasets. We further find that using our synthetic SVE gallery over the original real reference image provides a 22.5-point increase in recall, and a 10.4-point increase to AUC. And we find that \ours{} provides a 2.1-point increase over the headless DINOv3 using the SVE gallery, and a 1.6-point increase to AUC. This result demonstrates that the SVE gallery strategy provides a substantial boost over using a reference image, and the \ours{} head pushes this performance even further.

Across all evaluated backbones (DINOv3, DINOv2, and FLAIR), the SVE gallery consistently outperforms matching against a single reference image. Furthermore, equipping each backbone with our lightweight \ours{} projection head yields consistent gains over the SVE gallery baseline. These results establish two broad findings: first, SVE galleries provide a universally effective representation for category specification regardless of the underlying vision model; second, training a lightweight residual head over frozen backbone representations using \textsc{SynthAlikes} reliably amplifies the discriminative visual features necessary for precise dense categorization.

\paragraph{Number of SVEs}
\providecommand{\ph}{--}

\begin{table}[t]
  \centering
  \caption{%
    \textbf{Impact of Synthetic Visual Exemplar (SVE) Gallery Size on Performance.} The number of SVEs used in the positive and negative galleries are variable, and so we ablate over the number of SVEs used and it's impact on performance. We find performance saturates when the gallery size is restricted to 12 SVEs per-category. Evaluated using the \textsc{Lookalikes} dataset with oracle localizations.
  }
  \label{tab:sve_quantity}
  \setlength{\tabcolsep}{10pt}
  \begin{tabular}{cccc}
    \toprule
    Number of SVEs & Recall$\uparrow$ & FPR$\downarrow$ & AUC$\uparrow$ \\
    \midrule
    1  & 0.818 & 0.097 & 0.914 \\
    3  & 0.833 & 0.090 & 0.914 \\
    6  & 0.847 & 0.083 & 0.918 \\
    12 & 0.870 & 0.096 & 0.929 \\
    24 & 0.878 & 0.096 & 0.924 \\
    \bottomrule
  \end{tabular}
\end{table}
In Table~\ref{tab:sve_quantity}, we isolate the impact of gallery size by varying the number of SVEs per category while holding all other hyperparameters fixed. Notably, even a single SVE achieves strong performance, outperforming a single raw reference image (Table~\ref{tab:backbone_head_ablation}). This is surprising, but can be explained by two factors: real references are sourced automatically from repositories such as iNaturalist~\citep{inaturalist_2018, inaturalist} where images often feature isolated objects and variable quality and resolution, whereas our synthetic generation process reliably yields high-resolution scenes containing diverse, multi-instance depictions of the target class.

Furthermore, target category recall increases steadily with gallery size before largely saturating at $k{=}12$ SVEs, beyond which returns diminish marginally. Because this ablation evaluates categorization directly at ground-truth object locations, the observed gains reflect pure improvements in feature discrimination: scaling the gallery from $1$ to $12$ SVEs yields a $5.2$-point gain in recall, validating the benefit of multi-exemplar synthetic diversity.

\paragraph{SVE Pooling Operation}
\begin{table}[t]
  \centering
\caption{%
    \textbf{SVE Pooling Operation Ablation.} SVE galleries consist of synthetic images whose visual features originate from the underlying \ours{} module. Utilizing an SVE image as part of gallery for specifying a category necessitates a mechanism to pool its feature grid. We compare global average patch pooling against point-feature pooling via pseudo-detections, where point coordinates are extracted per image using an off-the-shelf counter (CountGD). These coordinates are indexed into the visual feature patch grid to extract interpolated point features. Evaluated using the \textsc{Lookalikes} dataset with oracle localizations.
  }
  \label{tab:pooling_op}
  \setlength{\tabcolsep}{8pt}
  \begin{tabular}{lccc}
    \toprule
    Operation &  Recall$\uparrow$ & FPR$\downarrow$ & AUC$\uparrow$ \\
    \midrule
    Whole-Image (GAP, all patches)  & 0.746 & 0.123 & 0.890 \\
    Per-Instance Pseudo-Detections & 0.825 & 0.098 & 0.916 \\
    Pooled Pseudo-Detection & \textbf{0.870} & \textbf{0.096} & \textbf{0.929} \\
    \bottomrule
  \end{tabular}
\end{table}

In Table~\ref{tab:pooling_op}, we evaluate strategies for pooling SVE vision feature grids into category representations. We find that global average pooling across all patches per SVE yields suboptimal results due to background contamination. Conversely, extracting interpolated point features at pseudo-detection coordinates (here, using the output of CountGD) focuses the category descriptors directly on target objects. This localized pooling strategy yields a performance boost, showing that point-based feature extraction is a necessary step for using SVE galleries. We further compare per-instance pseudo-detections to pooled pseudo-detections, where the per-instance strategy treats each detection as an independent exemplar, while the pooled strategy averages features across all detections in a single image. We find that pooling pseudo-detections leads to superior performance. Individual detections may occasionally experience background leakage or contain visual artifacts that degrade performance; averaging over the detected points extracts the dominant object signal from that SVE and provides significantly better recall.

\providecommand{\ph}{--}
\providecommand{\cmark}{\checkmark}
\providecommand{\xmark}{$\times$}

\begin{table}[t]
  \centering
\caption{%
    \textbf{Evaluating Exemplar Source and Quality.}
    We evaluate the performance gap between real and synthetic visual exemplars while holding all other parameters fixed; only the per-category exemplar gallery varies. To isolate the synthetic-to-real domain gap, we also compare synthetic exemplars containing a single canonical object (no crowding) against synthetic exemplars specifically prompted to contain crowds. Evaluated using the \textsc{Lookalikes} dataset with oracle localizations.
  }
  \label{tab:exemplar_source}
  \setlength{\tabcolsep}{8pt}
  \begin{tabular}{ccccc}
    \toprule
    Visual Exemplar & Crowding & Recall$\uparrow$ & FPR$\downarrow$ & AUC$\uparrow$ \\
    \midrule
    Synth. & \xmark & 0.735 & \textbf{0.096} & 0.882 \\
    Synth. & \cmark & 0.870 & 0.098 & 0.929 \\
    Real & \cmark & \textbf{0.925} & 0.101 & \textbf{0.930} \\
    \bottomrule
  \end{tabular}
\end{table}

\paragraph{Quality of Synthetic Exemplars.} Because our framework relies on synthetic exemplars, quantifying their downstream quality is important. In Table~\ref{tab:exemplar_source}, we compare our synthetic visual exemplars against 12 hand-curated real exemplars per category. Real exemplars yield equivalent AUC to synthetic ones, while providing a 5.5-point improvement in recall. We further compare two synthetic generation strategies: prompting for crowded, multi-object scenes versus single, canonical object depictions. Generating crowded scenes yields a 13.5-point improvement in recall and a 4.7-point improvement in AUC, showing that the SVE galleries benefit substantially from prompting with dense synthetic context.

\paragraph{In-Distribution vs. Zero-Shot Exemplars.} Table~\ref{tab:domain_adaptation_clean} evaluates \ours{} in the standard few-shot setting on the PairTally dataset using in-distribution visual exemplars extracted from the target domain. Compared to our zero-shot baseline, which relies strictly on an external reference, using an in-distribution exemplar reduces MAE by 5.06 points. This is important, because it demonstrates that despite our synthetic visual exemplars yielding superior performance in several settings, there is still a performance gap between the synthetic images and real in-distribution visual exemplars selected from the query image. 

\providecommand{\ph}{--}
\providecommand{\cmark}{\checkmark}
\providecommand{\xmark}{$\times$}

\begin{table}[t]
  \centering
  \caption{%
    \textbf{Effect of Reference Usage During Training and Inference.} Ablation evaluating performance across different combinations of train- and test-time references to assess the impact of reference guidance. Evaluated using the \textsc{Lookalikes} dataset with oracle localizations.}
  \label{tab:train_infer_ablation}
  \setlength{\tabcolsep}{8pt}
  \begin{tabular}{ccccc}
    \toprule
    \multicolumn{2}{c}{Reference Used} & & & \\
    \cmidrule(lr){1-2}
    Train & Infer & Recall$\uparrow$ & FPR$\downarrow$ & AUC$\uparrow$ \\
    \midrule
    \xmark & \xmark & 0.727 & 0.121 & 0.887 \\
    \cmark & \xmark & 0.807 & 0.127 & 0.903 \\
    \cmark & \cmark & \textbf{0.870} & \textbf{0.096} & \textbf{0.929} \\
    \bottomrule
  \end{tabular}
\end{table}
\paragraph{Reference Utilization.} In Table~\ref{tab:train_infer_ablation}, we quantify the impact of using real image references to guide SVE generation. First, holding the training setup fixed (trained with reference-guided SVEs), we perform inference using SVE galleries generated solely from text descriptions. Relying on real image references during generation rather than text alone yields a 6.3-point boost in recall and a 2.6-point improvement in AUC. Completely removing real image references from both training and inference leads to a dramatic 14.3-point drop in recall. These results demonstrate that real reference image guidance is important for downstream performance, as it grounds synthetic exemplars in fine-grained visual details that text prompts alone cannot capture.

Overall, our ablations validate our core methodology, and demonstrate that conditioning generation on real visual references, prompting for dense target scenes, and pooling features at pseudo-detection coordinates jointly yield an effective pipeline for zero-shot category specification using synthetic visual exemplars.
\subsection{Spatial Dynamics}
\paragraph{Crop-and-Reprocess}
\begin{table}[t]
  \centering
\caption{%
    \textbf{Benefits of Crop-and-Reprocess.} We evaluate the performance of our model when dense regions are adaptively cropped and then passed back through the model. We speculate that contamination from neighboring points impacts the separability of a target point. To evaluate this, we also report contaminated recall, which takes the original 32\texttimes 32 grid from DINOv3, and calculates recall for all target points that have a distractor sibling within a 3\texttimes 3 grid around that point. Evaluated using the \textsc{Lookalikes} dataset with oracle localizations.
  }
  \label{tab:crop_reprocess}
  \setlength{\tabcolsep}{8pt}
  \begin{tabular}{lccccc}
    \toprule
    Strategy  & Recall$\uparrow$ & FPR$\downarrow$ & AUC$\uparrow$ & Recall-Contam. $\uparrow$& FPR-Contam. $\downarrow$\\
    \midrule
    Base & 0.870 & \textbf{0.096} & 0.929 & 0.620  & \textbf{0.329}\\
    Crop  & \textbf{0.883} & 0.099 & \textbf{0.949} & \textbf{0.753} & 0.368\\
    \midrule
    $\Delta$  & +0.013 & +0.003 & +0.020 & +0.133 & +0.039\\
    \bottomrule
  \end{tabular}
\end{table}

Partitioning dense scenes and re-evaluating cropped regions has proven beneficial in prior counting architectures~\citep{xiong2023open,d2024afreeca}. Because our framework relies on a Vision Transformer with a $32 \times 32$ feature grid, and fine-grained datasets like \textsc{LookAlikes} and PairTally contain mixed-category scenes, objects from visually similar categories will fall within the same or adjacent patches. To mitigate this spatial aliasing, we identify dense regions where the base model detects multiple objects per patch, crop those areas, and adaptively rescale them so target objects occupy individual patches.

As shown in Table~\ref{tab:crop_reprocess}, evaluating our model with this adaptive crop-and-reprocess strategy (Section~\ref{sec:crop_and_reprocess}) yields consistent gains in overall recall and AUC. We also report \textit{contaminated recall}, which is defined as the recall measured strictly over query points that have a distractor sibling within a $3 \times 3$ patch neighborhood. On this sub-population, applying crop-and-reprocess achieves a 13.3-point improvement to contaminated recall, demonstrating its effectiveness at resolving spatial ambiguities in dense regions.

\providecommand{\ph}{--}

\begin{table}[t]
  \centering
  \caption{%
    \textbf{Spatial contamination.}
    We evaluate recall as a function of Chebyshev distance $d$ (in grid cells) to the nearest distractor object. We show that dense categorization breaks down when distractor objects share a patch with the target object. However, there also appears to be a degradation in performance when adjacent distractor objects multiple patches away. Evaluated using the \textsc{Lookalikes} dataset with oracle localizations.
  }
  \label{tab:spatial_contamination}
  \setlength{\tabcolsep}{8pt}
  \begin{tabular}{cc r ccc}
    \toprule
    \multirow{2}{*}{$d$} & \multirow{2}{*}{Neighborhood} & \multirow{2}{*}{$N$~~} & \multicolumn{3}{c}{Recall$\uparrow$} \\
    \cmidrule(lr){4-6}
    & & & Base & Crop & $\Delta$ \\
    \midrule
    0           & Same patch ($1\times 1$) &   636 & 0.5466 & 0.733 & $+0.169$ \\
    1           & $3\times 3$ shell        & 4,055 & 0.717 & 0.818 & $+0.101$ \\
    2           & $5\times 5$ shell        & 3,905 & 0.844 & 0.886 & $+0.042$ \\
    3           & $7\times 7$ shell        & 3,000 & 0.891 & 0.914 & $+0.023$ \\
    4           & $9\times 9$ shell        & 2,445 & 0.909 & 0.928 & $+0.019$ \\
    5\text{--}7 & Mid-range shells         & 5,143 & 0.920 & 0.934 & $+0.013$ \\
    $\ge 8$     & Far (same image)         & 5,997 & 0.935 & 0.939 & $+0.004$ \\
    \midrule
    $\infty$    & Single-cat (no sibling)  & 4,522 & 0.987 & 0.992 & $+0.005$ \\
    \bottomrule
  \end{tabular}
\end{table}
\paragraph{Spatial contamination.}
Table~\ref{tab:spatial_contamination} analyzes recall on the \textsc{LookAlikes} dataset as a function of the Chebyshev distance $d$ (in grid cells) to the nearest co-present lookalike distractor, evaluating points within concentric Chebyshev shells of radius $d$. This measures spatial contamination from neighboring related categories. As expected, the base filter recovers only $56.6\%$ of targets sharing a grid cell ($d{=}0$) with a distractor, where features are inherently entangled. However, recall improves monotonically as spatial separation increases, pointing to an additional neighborhood effect. We see that the recall difference between a single-category image and objects with $d{=}4$ is 7.8\%, even though the nearest object is 4 patches away. Because feature representations are extracted from a frozen vision backbone whose self-attention blends neighboring patch contexts, a distractor within a target's receptive field can corrupt the target descriptor before matching occurs, without the two objects landing in the same patch. This receptive-field leakage is apparently local, largely dissipating by $d{\geq}8$. This presents an additional explanation for why our crop-and-reprocess strategy directly counteracts this degradation: increasing the distance between objects weakens their interactions within the vision backbone.
\section{Conclusions}
We introduced \ours{}, a framework that decouples dense localization from dense categorization and specifies a category not with words but with images. From a single real reference image, our synthesis pipeline generates galleries of synthetic visual exemplars (SVEs) and the large-scale \textsc{SynthAlikes} corpus, turning open-world categorization into a matching problem against generated exemplars. Applied to a frozen counter, this image-guided zero-shot approach sets a new state of the art on the \textsc{LookAlikes}, PairTally, and PrACo benchmarks, with no annotation on the query image and no retraining of the counter. We further show that these gains are not tied to any single representation: training our lightweight discrimination head on \textsc{SynthAlikes} improves categorization across several frozen backbones, indicating that the synthetic supervision, rather than a particular feature extractor, is what surfaces the discriminative signal. By generating the visual specification a counter needs, \ours{} offers a scalable path to precise counting in the cluttered, look-alike-rich scenes where language alone falls short.


\bibliographystyle{recount}
\bibliography{main}


\end{document}